\documentclass[sigconf]{acmart}
\usepackage{multirow}
\usepackage{amsthm}
\usepackage{booktabs}
\usepackage{algorithm}
\usepackage{enumitem}
\usepackage{algorithmic}
\usepackage{bm}
\usepackage{color}
\usepackage{setspace}
\usepackage{float} 
\usepackage{subcaption}
\usepackage{soul}
\usepackage{graphicx}
\usepackage{balance}
\newcommand{\etal}{\textit{et al.}}

\usepackage{etoolbox}
\makeatletter
\patchcmd{\maketitle}{\@copyrightpermission}{
   \begin{minipage}{0.3\columnwidth}
     \href{https://creativecommons.org/licenses/by-nc-nd/4.0/}{\includegraphics[width=0.90\textwidth]{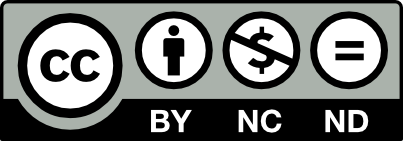}}
   \end{minipage}\hfill
   \begin{minipage}{0.7\columnwidth}
     \href{https://creativecommons.org/licenses/by-nc-nd/4.0/}{This work is licensed under a Creative Commons Attribution-NonCommercial-NoDerivs International 4.0 License.}
   \end{minipage}

   \vspace{5pt}
}{}{}

\makeatother

\AtBeginDocument{%
  \providecommand\BibTeX{{%
    \normalfont B\kern-0.5em{\scshape i\kern-0.25em b}\kern-0.8em\TeX}}}

\copyrightyear{2022}
\acmYear{2022}
\setcopyright{rightsretained}

\acmConference[MM '22]{Proceedings of the 30th ACM International Conference on Multimedia}{October 10--14, 2022}{Lisboa, Portugal} 
\acmBooktitle{Proceedings of the 30th ACM International Conference on Multimedia (MM '22), Oct. 10--14, 2022, Lisboa, Portugal} 
\acmISBN{978-1-4503-9203-7/22/10} \acmDOI{10.1145/3503161.3547943}

\begin{document}

\title{GSRFormer: Grounded Situation Recognition Transformer with Alternate Semantic Attention Refinement}
\author{Zhi-Qi Cheng}
\affiliation{
\institution{Carnegie Mellon University}
\country{}
}
\email{zhiqic@cs.cmu.edu}

\author{Qi Dai}
\affiliation{
\institution{Microsoft Research}
\country{}
}
\email{qid@microsoft.com}

\author{Siyao Li}
\affiliation{
\institution{Carnegie Mellon University}
\country{}
}
\email{siyaol@andrew.cmu.edu}

\author{Teruko Mitamura}
\affiliation{
\institution{Carnegie Mellon University}
\country{}
}
\email{teruko@cs.cmu.edu}

\author{Alexander G. Hauptmann}
\affiliation{
\institution{Carnegie Mellon University}
\country{}
}
\email{Alex@cs.cmu.edu}

\renewcommand{\shortauthors}{Zhi-Qi Cheng et al.}

\begin{abstract}
Grounded Situation Recognition (GSR) aims to generate structured semantic summaries of images for ``human-like'' event understanding.
Specifically, GSR task not only detects the salient activity verb (e.g. \textit{buying}), but also predicts all corresponding semantic roles (e.g. \textit{agent} and \textit{goods}).
Inspired by object detection and image captioning tasks, existing methods typically employ a two-stage framework: 1) detect the activity verb, and then 2) predict semantic roles based on the detected verb.
Obviously, this illogical framework constitutes a huge obstacle to semantic understanding.
First, pre-detecting verbs solely without semantic roles inevitably fail to distinguish many similar daily activities (e.g., \textit{offering} and \textit{giving}, \textit{buying} and \textit{selling}).
Second, predicting semantic roles in a closed auto-regressive manner can hardly exploit the semantic relations among the verb and roles.
To this end, in this paper we propose a novel two-stage framework that focuses on utilizing such bidirectional relations within verbs and roles.
In the first stage, instead of pre-detecting the verb, we postpone the detection step and assume a pseudo label, where an intermediate representation for each corresponding semantic role is learned from images.
In the second stage, we exploit transformer layers to unearth the potential semantic relations within both verbs and semantic roles.
With the help of a set of support images, an alternate learning scheme is designed to simultaneously optimize the results: update the verb using nouns corresponding to the image, and update nouns using verbs from support images.
Extensive experimental results on challenging SWiG benchmarks show that our renovated framework outperforms other state-of-the-art methods under various metrics\footnote{Code is available at \url{https://github.com/zhiqic/GSRFormer}}.
\end{abstract}

\maketitle

\vspace{-2mm}
\section{Introduction}
\label{sec:introduction}
\begin{figure}[!t]
  \centering
\vspace{3mm}
\includegraphics[width=0.85\linewidth]{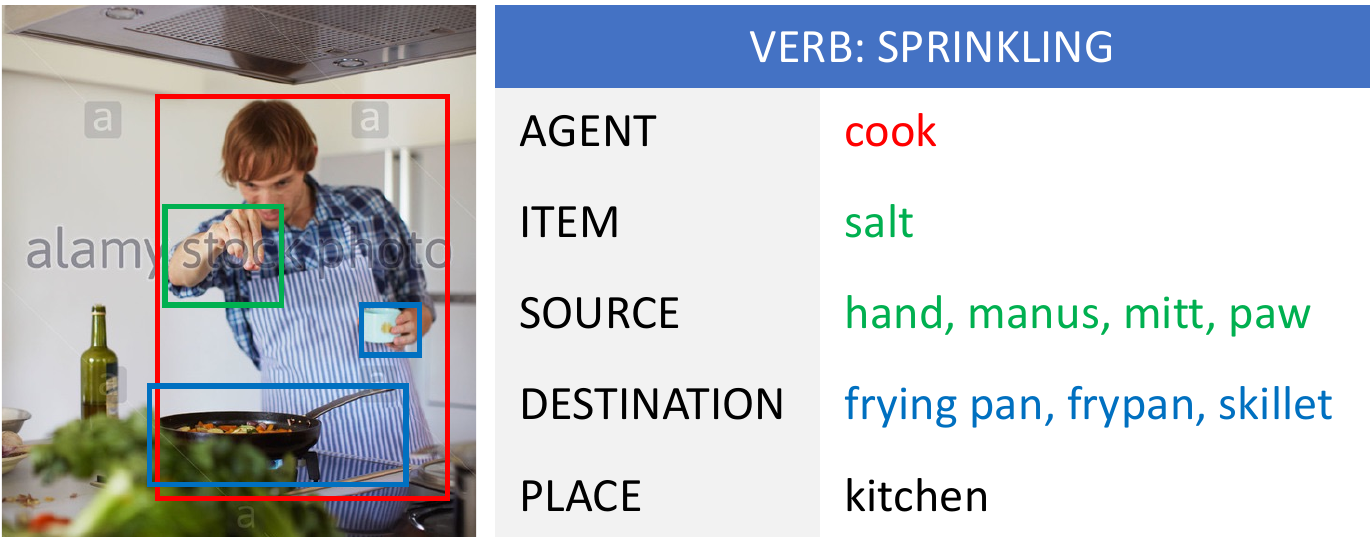}
\vspace{-3mm}
\caption{\small
Given an input image, Grounded Situation Recognition (GSR) not only detects the salient verb category (e.g., \textit{sprinkling}), but also predicts all corresponding semantic roles for sprinkling, such as \textit{agent: \textcolor{red}{man}}, {\textit{item}: \textcolor{green}{spice}}, and \textit{source: \textcolor{blue}{cup}},
etc.}
\label{fig:introduction}
\vspace{-6mm}
\end{figure}

Understanding complex events in a way that obeys human cognitive habits is one of the core tasks of computer vision and multimedia.
As shown in Figure \ref{fig:introduction}, ``human-like'' event understanding goes beyond traditional object- and action-centric detection and recognition tasks \cite{chen2021robust,dai2021dynamic,Liao_2020_CVPR,cheng2018learning,huang2018gnas,he2021db}.
Different from image captioning \cite{chen2017sca,nguyen2017vireo,zhao2021understanding,ji2021improving} and scene graph generation \cite{yang2018graph,xu2017scene,cong2021spatial,xiao2021video}, which use natural language or object-relational graphs to describe scenes, human-friendly event understanding must be event-centric, that is, not only to identify what activities happened, but also to recognize how objects participate in activities, i.e., answer questions like ``\textit{who is doing what with some tools at someplace}.''

\begin{figure*}[!t]
  \centering
  \small
  \includegraphics[width=0.9\linewidth]{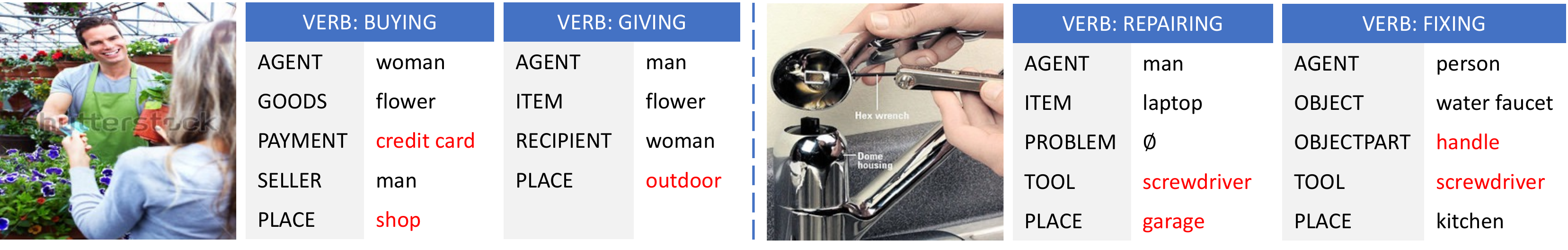}
  \vspace{-3mm}
  \caption{\small Some examples illustrate the importance of semantic relations in GSR tasks. Semantic relations are bidirectionally constrained, i.e., (left) noun entities can distinguish similar activities (verbs), and (right) similar verbs can control the occurrence of semantic roles (nouns).}
  \label{fig:sematic-example}
\vspace{-4mm}
\end{figure*}

To meet the demands of ``human-like'' event understanding, inspired by previous research on semantic role labeling \cite{palmer2010semantic,strubell2018linguistically,tan2018deep,conia2021unifying} in text, Grounded Situation Recognition (GSR) 
\cite{yatskar2016situation,pratt2020grounded} is proposed for event understanding in the multimedia domain.
As illustrated in Figure \ref{fig:introduction}, GSR not only detects the salient activity (verb) in the image (e.g., \textit{sprinkling}), but also recognizes all semantic roles (e.g., \textit{agent is man}, \textit{source is cup}).
To further determine semantic roles in images, GSR also provides visually grounded information (i.e., bounding boxes) for noun entities.
From the perspective of the theory of frame semantics \cite{fillmore2001frame}, GSR task can be considered as a multimedia extension to earlier lexical databases such as FrameNet \cite{baker1998berkeley} and PropBank \cite{kingsbury2002treebank}.
By describing activities with verbs and grounded semantic roles, GSR can provide a visually grounded verb-frame, which benefits many downstream scene understanding tasks, such as
information retrieval \cite{cheng2017selection,noh2017large,cheng2017video2shop,sun2021lightningdot},
question answering
\cite{anderson2018bottom,li2019relation,chen2020counterfactual,xiao2021video},
recommended system
\cite{chen2016context,cheng2016video,cheng2017video,sun2018personalized},
and multimedia understanding \cite{zhao2021understanding,sadhu2021visual,hao2022group,zhang2021token}.

\begin{figure*}[!t]
  \centering
  \includegraphics[width=0.9\linewidth]{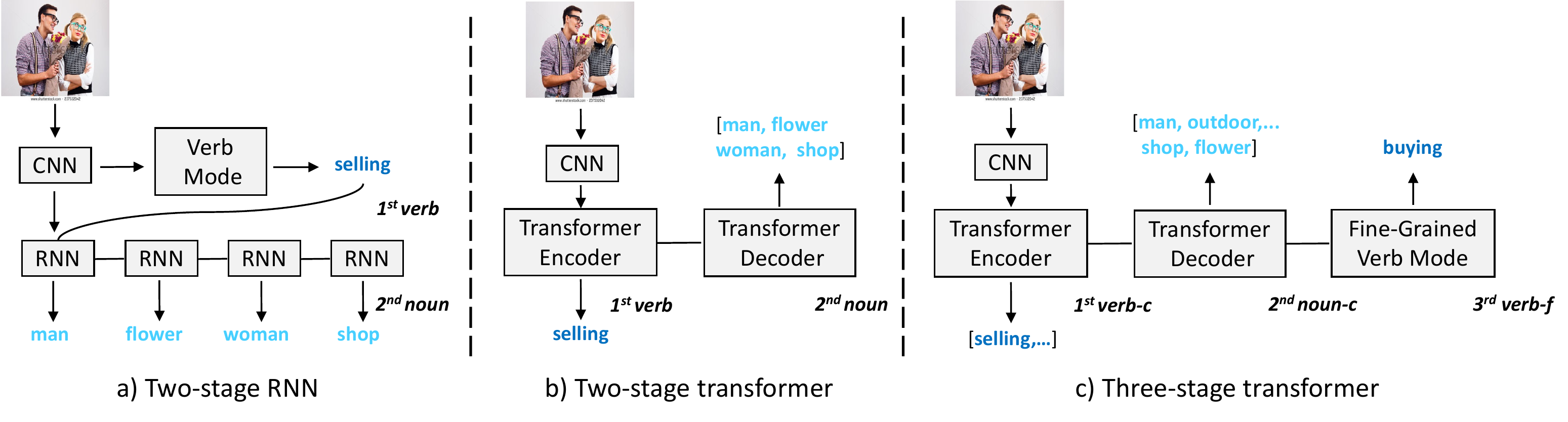}
  \vspace{-4mm}
  \caption{\small GSR task learning framework.
(a) Two-stage RNN pre-detects verbs in the first stage and predicts noun entities in an auto-regressive manner in the second stage.
(b) Two-stage transformer first uses the encoder to detect verbs and then uses the decoder to predict noun entities.
(c) Three-stage transformer \cite{wei2021rethinking,cho2022CoFormer} adopts a coarse-to-fine refinement process.
It still obeys the two-stage idea, where the first stage detects a set of similar verbs and the second stage recognizes noun entities.
The third stage utilizes noun entities to refine verbs.}
  \label{fig:framework-problem}
\vspace{-3mm}
\end{figure*}

To sum up, the essence of GSR task is using semantic relations to generate verb-frames for event understanding.
However, inspired by object detection and image captioning, almost all current GSR methods \cite{mallya2017recurrent, cho2021grounded,li2017situation,suhail2019mixture,pratt2020grounded} adopt a two-stage framework.
As shown in Figure \ref{fig:framework-problem} (a-b), two-stage framework
1) first blindly pre-detects verbs to reduce the search space, and then 
2) predicts the semantic roles in an auto-regressive (RNN) or parallel (Transformer) manner depending on the detected verbs.
Such two-stage frameworks obviously neglect the semantic relations among verbs and semantic roles.
On the one hand, pre-detecting the verb without noun entities inevitably fails to distinguish some similar daily activities.
For example, as shown in Figure \ref{fig:sematic-example} (left), it is hard to distinguish similar verbs (e.g., \textit{buying} and \textit{giving}) without the help of any noun entities.
On the other hand, based on pre-identified verbs, applying auto-regression in a closed space would accumulate errors and thus miss the semantic relationships.
As shown in Figure \ref{fig:sematic-example}, once verb \textit{buying} is wrongly predicted as \textit{giving} in the first stage, the semantic roles \textit{payment: credit} and \textit{place: shop} could be neglected in the second stage.

Similar to our starting point, previous work CoFormer \cite{cho2022CoFormer} and SituFormer \cite{wei2021rethinking} also argue that the existing two-stage framework is problematic.
As shown in Figure \ref{fig:framework-problem} (c), a three-stage framework is thus proposed to further optimize verbs with predicted nouns.
Following traditional two-stage works, the first two stages perform the verb and noun entities detection, respectively.
Then, in the third stage, the predicted noun entities are used to refine the verb.
However, such disentangled framework still has the following flaws.
First, the bidirectional semantic relations between verbs and noun entities cannot be fully exploited.
They either use the noun roles to refine the verbs (SituFormer), or only use verbs to refine the noun roles (CoFormer) while ignoring each other.
Second, the framework is redundant.
It has two parallel transformer verb and noun detectors, but does not learn semantic relations between them during the encoding phase.
Third, the refinement process is not scalable.
It can be treated as a one-time noun-to-verb optimization, which apparently cannot be expanded.

To address these issues, we focus on how to exploit such bidirectional semantic relations within verbs and noun entities, which can constrain each other.
Rather than explicitly pre-detecting the verbs at the first stage, we postpone decision verbs, thus simply assuming a pseudo-category and learning an intermediate representation for each noun entity.
We then devise an iterative framework to capture the semantic relations among verbs and nouns and alternately learn their features.
As such, we streamline the redundant structures and make them flexibly handle semantic roles in parallel.

Technically, our proposed method, called GSRFormer, is built based on the transformer structure \cite{vaswani2017attention} due to its parallel processing capability.
As shown in Figure \ref{fig:framework-solution}, GSRFormer adopts a two-stage architecture that consists of an encoder and a decoder.
In the encoder part, we first utilize stacked Multi-Head Attention (MHA) layers to learn the feature of the verb.
By assuming a pseudo category of the verb, we further learn the intermediate representations of the corresponding noun entities from the image.
In the decoder part, MHA layers are employed to mine the implicit relations among both verbs and nouns. By leveraging a set of support images, the model alternately optimizes the verbs and nouns in a loop: update the features of verbs using nouns, and vice versa.
Our framework successfully learns the semantically rich representations for both verbs and nouns and thus performs accurate recognition.
To conclude, our contributions are mainly three folds:

\begin{itemize}[itemsep=2pt,topsep=2pt, parsep=2pt]
\setlength{\topsep}{1mm}
\setlength{\itemsep}{1mm}
\item We reveal the problems of existing frameworks and point out that learning the bidirectional semantic relations is the core for accurate role recognition.
\item We propose a two-stage framework with transformer structures to iteratively refine activity verbs and noun entities. It flexibly mines the potentially open semantic relations within verbs and nouns and alternately updates their features.
\item Extensive experiments on challenging SWiG benchmarks fully demonstrate that our proposed framework outperforms other state-of-the-art methods under various metrics.
\end{itemize}
\vspace{-2mm}

\vspace{-3mm}
\section{Related Work}
\label{sec:related-work}
\noindent \textbf{History of SR to GSR tasks}.
Although deep learning achieves satisfactory performance in basic vision tasks such as action recognition~\cite{wang2013action,yang2020temporal,tran2018closer,zhao2017single,carreira2017quo,he2021db,Liao_2020_CVPR} and object detection~\cite{redmon2016you, lin2017feature, tan2020efficientdet,carion2020end,chen2021robust,dai2021dynamic}, they still cannot fully understand events in natural scenes.
To address this problem, image captioning~\cite{stefanini2022show,you2016image,rennie2017self,cornia2020meshed,anderson2018bottom,xu2021towards} and scene graph generation~\cite{li2017scene,xu2017scene,yang2018graph,zellers2018neural,Chen_2019_ICCV,cong2021spatial,wei2020hose} attempt to reason and describe scene content through natural language or relational graphs of objects.
However, these efforts have still failed to understand events consistent with human cognition, i.e., identifying what happened and who was involved in what roles.

In this context, Yatskar~\etal~\cite{yatskar2016situation} first proposed Situation Recognition (SR) task and annotated imSitu dataset as the benchmark.
However, the original SR task cannot point to where the involved noun entities are located in the image.
To further address the visual grounding of the entities, Pratt \etal~\cite{pratt2020grounded} redefines Grounded Situation Recognition (GSR) task and proposes SWiG dataset by providing bounding box annotations on imSitu dataset.
GSR task can be seen as a further extension of SR task.

\noindent \textbf{Challenges of GSR tasks}.
The challenges of GSR task are to handle the semantic relations in the scene.
Yatskar~\etal~\cite{yatskar2016situation} proposed a conditional random field (CRF) model in the initial phase.
However, follow-up work~\cite{yatskar2017commonly} pointed out that CRF method cannot effectively utilize semantic relations.
Since then, a lot of works have started to investigate how to model the relations among semantic roles.
The previous technical routes mainly include Recurrent Neural Network (RNN)~\cite{mallya2017recurrent,pratt2020grounded,vicol2018moviegraphs,vicol2018moviegraphs}, Graph Neural Network (GNN)~\cite{li2017situation,suhail2019mixture} and Relational Reasoning~\cite{cooray2020attention,cadene2019murel}, etc.

\noindent \textbf{Transformers in GSR task}.
After great success in NLP tasks, transformer structure~\cite{vaswani2017attention} was introduced to solve various computer vision problems, including image generation \cite{parmar2018image,child2019generating}, image recognition \cite{dosovitskiy2020image,liu2021swin,touvron2021training},
object detection \cite{carion2020end,zhu2020deformable},
object segmentation \cite{ye2019cross},
image captioning \cite{cornia2020meshed} and scene graph generation \cite{cong2021spatial}, etc.
To further exploit the strength of the transformer in GSR task, Cho~\etal~\cite{cho2021grounded} proposed the first transformer framework (GSRTR) by replacing the object queries in transformer object detector (DETR \cite{zhu2020deformable}) with semantic role queries.
Wei~\etal~\cite{wei2021rethinking} recently proposed SituFormer, which uses two transformer-based verb and noun detectors to improve the performance.
In addition, Cho~\etal~\cite{cho2022CoFormer} proposed CoFormer that tends to leverage the semantic relations between the verb and noun roles to refine the prediction.

\noindent \textbf{Problem of framework}.
To elaborate, as shown in Figure \ref{fig:framework-problem}, almost all existing GSR methods \cite{mallya2017recurrent,pratt2020grounded,vicol2018moviegraphs,suhail2019mixture, cooray2020attention,cadene2019murel,cho2021grounded} adopt a two-stage framework based on RNN or transformer.
In RNN structure, Pratt \etal~\cite{pratt2020grounded} proposes a Joint Situation Localizer (JSL) model, which consists of a verb classifier in the first stage and an RNN-based object detector in the second stage.
In transformer structure, GSRTR~\cite{cho2021grounded} uses a transformer encoder to detect verbs in the first stage and a transformer decoder to predict semantic roles in the second stage.
These two-stage frameworks evidently cannot exploit the semantic relations.
First, recklessly detecting verbs in the first stage will inevitably misidentify similar activities.
Second, the closed auto-regressive strategy of the second stage not only fails to correct misrecognized verbs, but leads to more mispredicted semantic roles.
Although recent works (SituFormer \cite{wei2021rethinking} and CoFormer \cite{cho2022CoFormer}) are also aware of framework issues and adopt a three-stage framework to refine prediction results from coarse to fine, they are still unable to optimize the results with bidirectional semantic relationships (i.e., from both verbs and nouns).

\section{Proposed Method}
\label{sec:proposed-method}
\subsection{Problem Formulation}
\label{sec:problem-definition}
\noindent {\textbf{Definition of GSR task}}.
Given an image $I$, GSR aims to generate a structured verb frame $\mathcal{F}_v = \left\{v, \mathcal{R}_v\right\}$\footnote{These predefined verb frames are filtered from PropBank~\cite{kingsbury2002treebank} or FrameNet~\cite{fillmore2003background,baker1998berkeley}.}.
As shown in Figure \ref{fig:introduction}, GSR not only recognizes the salient verb $v \in \mathcal V $, but also detects all corresponding semantic roles $\mathcal R_v=\{ \left(r, n_r, \mathbf{b}_r \right) \vert\; \mathrm{for} \; r\in\mathscr R_v\}$, where $\mathscr R_v = \{r_1,...,r_m\}$ is the set of role types for verb $v$.
For instance, the verb-frame in Figure \ref{fig:introduction} can be instantiated as
$\mathcal{F}_v = \big(Sprinkling,$
$\big\{(Agent,$ $Man,$ $\textcolor{red}{\Box}),$
$(Item,$ $Spice,$ $\textcolor{green}{\Box}),$
$(Source,$ $Cup,$ $\textcolor{blue}{\Box}),$
$(Destination,$ $Pan,$ $\textcolor{blue}{\Box}),$
$(Place,$ $Kitchen,$ $\emptyset_b)\big\}\big)$.
Semantic roles $\mathcal R_v$ is a collection of triples, where each role contains the role type $r$, the noun entity $n_r \in \mathcal N$ and the corresponding bounding box $\mathbf b_r \in \mathbb R^4$.
Note that not all semantic roles have corresponding nouns and bounding boxes, i.e., $n_r$ or $\mathbf b_r$ can be equal to $\left\{ \emptyset \right\}$.

\begin{figure*}[!t]
  \centering
  \includegraphics[width=0.9\linewidth]{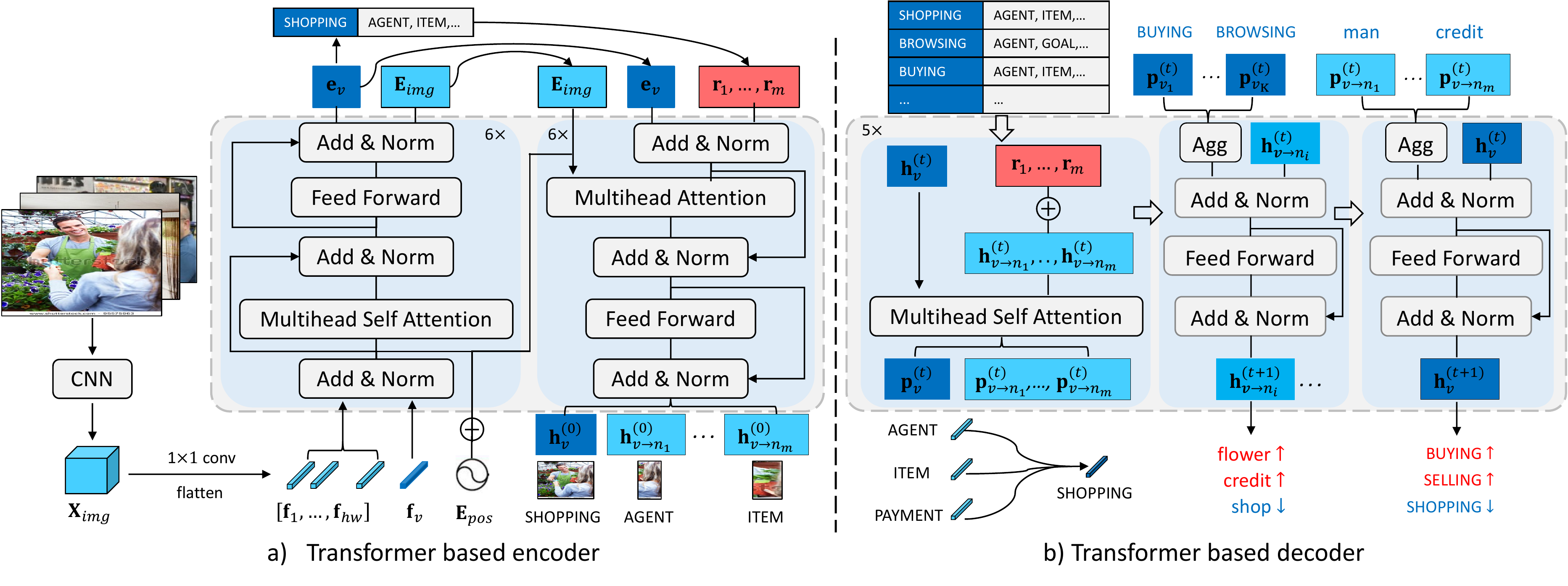}
  \vspace{-5mm}
 \caption{\small GSRFormer follows the classic encoding and decoding framework.
Transformer encoder utilizes two multi-head attention modules to learn the intermediate features for verbs and semantic roles.
Transformer decoder consists of four parts.
1) A set of similar activities (verbs) are retrieved using features from the encoder.
2) A multi-head attention layer is used to compute the messages $\mathbf{p}$, thus capturing semantic relations among verbs and nouns.
3-4) The features of nouns and verbs are updated alternately with the computed messages.
Note that we stack multiple decoder layers for iterative optimization (refinement).
}
\vspace{-4mm}
  \label{fig:framework-solution}
\end{figure*}

\noindent \textbf{Problems of existing frameworks}.
Inspired by object detection and image captioning, as shown in Figure \ref{fig:framework-problem}, the existing GSR methods \cite{mallya2017recurrent, cho2021grounded,li2017situation,suhail2019mixture,pratt2020grounded} widely adopt a two-stage framework, i.e.,
1) identifying salient verbs $v$ in the first stage, and 
2) detecting the corresponding semantic roles $\mathcal R_v$ in the second stage:
\begin{equation}
\mathcal P(\mathcal F_v | I) = 
\underbrace{\mathcal P(v | I)}_{\text{verb}} 
\underbrace{\mathcal P(\mathcal R_v | v, I)}_{\text{noun}}.
\end{equation}
There are multiple highly similar activities in GSR tasks, such as \textit{buying} and \textit{giving}, as shown in Figure \ref{fig:sematic-example}.
This two-stage framework is inherently unable to utilize the noun entities in the first stage to distinguish similar verbs, let alone misidentify semantic roles due to the verb prediction errors accumulated in the second stage.

To address these issues, Wei \etal~\cite{wei2021rethinking} recently proposed a three-stage framework (SituFormer) as,
\begin{equation}
\mathcal P(\mathcal F_v | I) = 
\underbrace{\mathcal P(\mathcal V_c | I)}_{\text{verb-c}} 
\underbrace{\mathcal P(\{\mathcal{R}_v\}_c| \mathcal V_c, I  )}_{\text{noun-c}}
\underbrace{\mathcal P(v, \mathcal R_v | \mathcal V_c, \{\mathcal{R}_v\}_c, I)}_{\text{verb-f}},
\end{equation}
where the main idea is to refine verb predictions in a coarse-to-fine manner.
As shown in Figure \ref{fig:framework-problem} (c), it uses the first two stages to identify a set of candidate verbs $\mathcal V_c$ and corresponding noun entities $\{\mathcal{R}_v\}_c$.
Then the third stage uses a ranking loss to refine the verb $v$ with a set of support images through similarity retrieval.

Similarly, Cho \etal~\cite{cho2022CoFormer} also proposed a three-stage  framework (CoFormer).
From the framework perspective, CoFormer is similar to SituFormer.
1) The first stage coarsely predicts noun roles (Glance). 
2) The second stage uses the predicted noun roles to help predict the verb (Gaze-Step1). 
3) The third stage refines the candidate noun roles in Gaze-Step1 using the predicted verb (Gaze-Step2).

Obviously, these revised frameworks are still unreasonable due to the following issues.
1)~It uses only noun entities to refine the verb, but totally ignores semantic constraints in the opposite direction (i.e., from verbs to noun entities).
2)~Although it has two redundant transformer-based noun and verb detectors, it does not learn the semantic relations between them at the encoding stage.
3)~The coarse-to-fine refinement process can only be done once, and it is impossible to iteratively optimize both verbs and nouns.

\vspace{-2mm}
\subsection{Overview of GSRFormer}
\label{sec:overview-gsr-former}
To solve these problems, we reconstruct a framework (GSRFormer) with transformers to fully exploit the semantic relations of GSR tasks.
We reformulate GSR task as,
\begin{equation}
\mathcal P(\mathcal F_v | I)=
\underbrace{\mathcal P(\tilde{v}, \mathcal H_{v}| I)}_{\text{step-1: encoding}}
\underbrace{\mathcal P(v,\mathcal R_v|\mathcal H_{v}, I)}_{\text{step-2: decoding}},
\end{equation}
where $\tilde{v}$ is the assumed pseudo verb category, and $\mathcal H_{v}$ is the intermediate feature for the verb and associated roles.
The godsend is that it yields the essence of the GSR task, which is to extract comprehensive semantic relations and then iteratively refine the results.
As shown in Figure \ref{fig:framework-solution}, GSRFormer consists of a transformer-based encoder and decoder, respectively.
In the first step (Sec.~\ref{sec:tranformer-encoder}), instead of pre-detecting verbs, the transformer encoder learns intermediate representations $\mathcal H_v$ of verbs and corresponding semantic roles from images, respectively.
In the second step (Sec.~\ref{sec:transform-decoder}), we first use the features obtained in the encoder to retrieve the representations of Top-$K$ similar verbs as the support verb set $\{\mathcal H_v\}_s$.
To mine the various semantic relations, the neural message passing mechanism \cite{gilmer2017neural} is then used to flexibly associate relevant relations to each verb or role, thus effectively updating their representations.
Finally, we take full advantage of the transformer structure to perform multiple iterations to refine verbs $v$ and semantic roles $\mathcal R_v$.

\vspace{-2mm}
\subsection{Transformer-based Encoder}
\label{sec:tranformer-encoder}
The transformer encoder is devised to learn intermediate representations of verbs and semantic roles from images using two multi-head attention modules, respectively.

\noindent \textbf{Representation of verbs}.
As shown in Figure \ref{fig:framework-solution}, given an image $I$, CNN backbone first extracts the feature map
$\mathbf X_{\text{img}} \in \mathbb R^{c \times h \times w}$.
Since the input to the transformer encoder is a sequence of tokens, a $1\times 1$ convolutional layer and a flatten operator are used to convert the $\mathbf X_{\text{img}}$ into a sequence of ``visual'' tokens $[\mathbf f_1,..., \mathbf {f}_ {hw }]$, where each token $\mathbf f_i \in \mathbb R^d$ is compressed as a $d$-dim visual feature.
Inspired by the classification token used in ViT~\cite{dosovitskiy2020image}, we initialize a learnable verb token $\mathbf{f}_v \in \mathbb R^d$ to stands for verb $v$.
Then all visual and verb token sequences are fed into the first encoding module as,
\begin{equation}
\begin{aligned}
\left[\mathbf e_v, \mathbf e_1,..., \mathbf e_{hw}\right]=\\ 
\text{MHA}_{\text{img}}^{\text{verb}} (\left[\mathbf f_v, \mathbf f_1,..., \mathbf f_{hw}\right]\oplus\mathbf E_{pos}),
\end{aligned}
\end{equation}
where $\oplus$ is element-wise addition and $\mathbf E_{pos}$ is positional embedding used to distinguish relative positions in the sequence.
{\small $\text{MHA}_{\text{img}}^{\text{verb}}$} is a set of stacked multi-head self-attention blocks.
As shown in Figure \ref{fig:framework-solution}, each block consists of a multi-head self-attention layer and a feed-forward network, and the layer normalization \cite{ba2016layer} (Add \& Norm) is used before both of them\footnote{$\mathbf E_{pos}$ has the same dimension as sequences of visual and verb tokens. For more implementation details, see original papers \cite{vaswani2017attention,cho2021grounded}.}.
The role of {\small $\text{MHA}_{\text{img}}^{\text{verb}}$} module is to use multi-head self-attention to learn the intermediate representation of the verb from image features.
The output can be divided into optimized 1) image features ${\small \mathbf E_{\text{img}}=[\mathbf e_1,..., \mathbf e_{hw}] \in \mathbb R^{d \times hw}}$ and 2) verb feature ${\small \mathbf e_v \in \mathbb R^{d}}$.
Here verb feature $\mathbf e_v$ is fed to a classifier to determine a pseudo verb category $\tilde{v}$.
After obtaining $\tilde{v}$, we can fetch the corresponding semantic roles $\mathscr R_v$ and initialize them to a sequence ${\small [\mathbf r_1,..., \mathbf r_m]}$, where each initialized role ${\small \mathbf r_i \in \mathbb R^ d}$ is a $d$-dim visual feature.

\noindent \textbf{Representation of semantic roles}.
We further learn intermediate representations of corresponding semantic roles plus verb {\small $\mathcal H_v=[\mathbf h_{v}^{(0)}, \mathbf h_{v \to n_1}^{(0)}, ..., \mathbf h_{v \to n_m}^{(0)}]$} from images.
Specifically, ${\small \mathbf E_{\text{img}}}$ and ${\small \mathbf e_v}$ are used as the input to the second encoding module as,
\begin{equation}
\begin{aligned}
\left[\mathbf h_{v}^{(0)}, \mathbf h_{v \to n_1}^{(0)}, ..., \mathbf h_{v \to n_m}^{(0)}\right] =\\ \text{MHA}_{\text{img}}^{\text{roles}} (\underbrace{\mathbf E_{pos}\oplus \mathbf E_{img}}_{\text{key/value}},\underbrace{[\mathbf e_{v}, \mathbf r_1, ..., \mathbf r_m]}_{\text{query}}),
\end{aligned}
\end{equation}
where we concatenate verb feature $\mathbf e_{v}$ and corresponding semantic role embeddings ${\small [\mathbf r_1 ,..., \mathbf r_m]}$ as query.
As shown in Figure \ref{fig:framework-solution}, similar to the previous encoding module, {\small $\text{MHA}_{\text{img}}^{\text{roles}}$} is also a set of stacked multi-head attention blocks.
The output representation {$\mathcal H_v$} is then utilized in the decoder part.
After GSRFormer encoder, we assume that the feature $\mathcal H_v$ has captured the semantics from the image.

\subsection{Transformer-based Decoder}
\label{sec:transform-decoder}
The transformer decoder aims to utilize the semantic relations among a set of similar verbs and corresponding semantic roles to simultaneously refine their features.
While transformer itself is a powerful model that can mine the implicit relations of all pairwise interactions, it still struggles to leverage additional domain-specific knowledge.
In our framework, we expect to exploit a set of support verbs (excluding their corresponding nouns) to profit the noun refinement, while only using the nouns in one single image to refine its verb.
This \textit{a priori} knowledge can hardly be included when directly applying the transformer since it considers ALL pairwise relations.
To this end, we borrow ideas from the neural message passing \cite{gilmer2017neural}, where the ``messages'' are first computed by transformers, and then each verb or noun entity is updated using ``messages'' arising from the relations to other appropriate entities.
As shown on the right of Figure \ref{fig:framework-solution}, the transformer decoder is mainly composed of the following four parts.

\noindent \textbf{Support verbs set}.
There are many very similar verbs in GSR tasks, as shown in Figure \ref{fig:sematic-example}.
Inspired by previous work \cite{wei2021rethinking}, we use the features $\mathcal H_{v}= [\mathbf h_{v}, \mathbf h_{v \to n_1}, ..., \mathbf h_{v \to n_m}]$ learned from the encoder to retrieve the features of top-$K$ similar verbs $\{\mathcal H_v\}_s = \{\mathcal H_{v_1},...,\mathcal H_{v_K}\}$ as support verbs set,
\begin{equation}
\{\mathcal H_v\}_s = \mathop{\arg}_{\mathcal H_{v_j}}~~\mathop{\text{top-}K}_{v_j \in \mathcal{D}_{\mathcal{T}}} \;\text{S}(\mathcal H_v, \mathcal H_{v_j}),
\end{equation}
where $\mathcal{D}_{\mathcal{T}}$ is the set of all training images.
The similarity score $\text{S}(\cdot,\cdot)$ is the average cosine similarity of semantic roles as,
\begin{equation}
\text{S}(\mathcal H_v, \mathcal H_{v_j}) = \frac{1}{m} \sum_{i=1}^{m} \text{sim}(\mathbf h_{v \to n_i}^{(0)},\mathbf h_{v_j \to n_i}^{(0)}),
\end{equation}
where $\text{sim}(\cdot)$ is cosine similarity.
$\mathbf h_{v \to n_i}$ and $\mathbf h_{v_j \to n_i}$ are the noun entity features for the verbs $v$ and $v_j$, respectively.

\noindent\textbf{Semantic relation message computation}.
Given retrieved support set $\{\mathcal H_v\}_s$, we aim to update the feature representation of each element in $\mathcal H_v$ by leveraging the relations between or within $\mathcal H_v$ and $\{\mathcal H_v\}_s$.
Following the standard message passing paradigm, we compute one message $\mathbf{p}$ for each involved verb and noun in one image by a multi-head attention (MHA) layer as,
\begin{equation}
\begin{aligned}
\left[
        \mathbf{p}_{v}^{(t)},
        \mathbf{p}_{v \to n_1}^{(t)},
        \ldots,
        \mathbf{p}_{v \to n_{m}}^{(t)}
    \right]
    =\\
    \text{MHA}^{\text{roles}}_{\text{verb}}\left(
        \left[
            \mathbf{h}_{v}^{(t)},
            \mathbf r_1 \oplus \mathbf{h}_{v \to n_1}^{(t)},
            \ldots,
            \mathbf r_m \oplus \mathbf{h}_{v \to n_m}^{(t)}
        \right]
    \right),
\end{aligned}
\end{equation}
where $\oplus$ is element-wise addition, and $t$ implies the $t$-th iteration.
$\mathbf{h}_{v}^{(t)}$ and $\mathbf{h}_{v \to n_{(\cdot)}}^{(t)}$ are the learned verb and noun entity features, respectively.
Here we replace the positional encoding in the original MHA layer with the semantic role embedding $\mathbf r_{(\cdot)}$.
Thus {\small $\text{MHA}^{\text{roles}}_{\text{verb}}$} is actually a multi-head self-attention model.
We perform the relation message computation for the entire support set.
The obtained verb message $\mathbf{p}_{v}^{(t)}$ and the noun entity message $\mathbf{p}_{v\to n_{(\cdot)}}^{(t)}$ contain all the semantic information within a single image.
Below we will consider semantic relations in multiple verbs (i.e., support verbs set).

\noindent \textbf{Refine noun entity with verbs}.
We utilize the semantic relations (messages) from the verbs of support set to refine the noun entities.
To update the representation of a noun entity $\mathbf{h}_{v \to n_i}^{(t)}$, a single update message $\mathbf{p}_{v_{all}}^{(t)}$ is computed by aggregating the messages of support set verbs $\{\mathbf{p}_{v_1}^{(t)},...,\mathbf{p}_{v_K}^{(t)}\}$ as Eq. \ref{eq:agg-v}.
The aggregation function $\text{Agg}(\cdot)$ can be any permutation-invariant function (e.g., element-wise sum and max), and here we employ a gated update function \cite{li2016gated}.
After the messages are fused, a transformer sublayer (FFN and LN) are used to updated representation with residual connection:
\begin{align}
\label{eq:agg-v}
\mathbf{p}_{v_{all}}^{(t)} = \text{Agg}(
        \{
            \mathbf{p}_{v_k}^{(t)} \mid \mathrm{for} \; 1\leq k \leq K
        \}
    ),
\\
\mathbf{q}_{v \to n_i}^{(t)} = \text{LN}\left(\mathbf{h}_{v \to n_i}^{(t)} + \mathbf{p}_{v_{all}}^{(t)}\right),\\
\mathbf{h}_{v \to n_i}^{(t+1)} = \text{LN}\left(\mathbf{q}_{v \to n_i}^{(t)} + \text{FFN}\left(\mathbf{q}_{v\to n_i}^{(t)}\right)\right),
\end{align}
where $\text{LN}(\cdot)$ is layer normalization \cite{ba2016layer} and $\text{FFN}(\cdot)$ is a feed-forward neural network (commonly with one large intermediate layer).

\noindent \textbf{Refine verb with noun entities}.
Similarly, when updating the verb feature, we utilize the messages of nouns only from the single associated image, as shown in Eq. \ref{eq:agg-n}-\ref{eq:h-v-n},
\begin{align}
\label{eq:agg-n}
\mathbf{p}_{n_{all}}^{(t)} = \text{Agg}(
        \{
            \mathbf{p}_{v \to n_i}^{(t)} \mid \mathrm{for} \; 1 \leq i \leq m
        \}
    ),
\\
\label{eq:q-v-n}
\mathbf{q}_{v }^{(t)} = \text{LN}\left(\mathbf{h}_{v}^{(t)} + \mathbf{p}_{n_{all}}^{(t)}\right),
\\
\label{eq:h-v-n}
\mathbf{h}_{v}^{(t+1)} = \text{LN}\left(\mathbf{q}_{v}^{(t)} + \text{FFN}\left(\mathbf{q}_{v}^{(t)}\right)\right).
\end{align}
Note that the above two refining processes can be accomplished alternately.
Unlike the previous work \cite{wei2021rethinking}, which only optimized from rough noun entities to verbs, our framework takes full advantage of the flexibility of the transformer structure to efficiently perform multiple refinement iterations.

After completing the refinement of noun entities and verb features for $T$ iterations, we employ a lightweight MLP over the verb and noun entity features, respectively, to achieve the classification of verbs and the regression of noun entities with bounding boxes,
\begin{align}
{v} = \text{MLP} (\mathbf h_{v}^{T}),
\\
\{{n_i}, {\mathbf b_i} \} = \text{MLP} (\mathbf h_{v \to n_i}^{T}).
\end{align}
We discuss the training process in detail in the following section.
In the ablation studies, we discuss the effect of the number of loops and refinement order on the results in detail.

\begin{table*}[t]
\small
\caption{\small \label{tab:dev_set_result} Performance (\%) comparisons of GSRFormer (ours) and baseline methods on SWiG dataset development set.}
\vspace{-3mm}
\resizebox{1.97\columnwidth}{!}{
\begin{tabular}{l|ccccc|ccccc|cccc}
\hline
\multicolumn{1}{c|}{\multirow{2}{*}{Models}} & \multicolumn{5}{c|}{Top-1-Verb}                                                    & \multicolumn{5}{c|}{Top-5-Verb}                                                    & \multicolumn{4}{c}{Ground-Truth-Verb}              \\
\multicolumn{1}{c|}{}                        & verb           & value          & val-all        & grnd           & grnd-all       & verb           & value          & val-all        & grnd           & grnd-all       & value          & val-all        & grnd  & grnd-all \\ \hline
\hline
\multicolumn{15}{c}{Methods for Situation Recognition} \\  \hline  
CRF \cite{yatskar2016situation}                                         & 32.25          & 24.56          & 14.28          & –              & –              & 58.64          & 42.68          & 22.75          & –              & –              & 65.90          & 29.50          & –     & –        \\
CRF+DataAug \cite{yatskar2017commonly}                                 & 34.20          & 25.39          & 15.61          & –              & –              & 62.21          & 46.72          & 25.66          & –              & –              & 70.80          & 34.82          & –     & –        \\
VGG+RNN \cite{mallya2017recurrent}                                     & 36.11          & 27.74          & 16.60          & –              & –              & 63.11          & 47.09          & 26.48          & –              & –              & 70.48          & 35.56          & –     & –        \\
FC-Graph \cite{li2017situation}                                    & 36.93          & 27.52          & 19.15          & –              & –              & 61.80          & 45.23          & 29.98          & –              & –              & 68.89          & 41.07          & –     & –        \\
CAQ \cite{cooray2020attention}                                         & 37.96          & 30.15          & 18.58          & –              & –              & 64.99          & 50.30          & 29.17          & –              & –              & 73.62          & 38.71          & –     & –        \\
Kernel-Graph \cite{suhail2019mixture}                                & 43.21          & 35.18          & 19.46          & –              & –              & 68.5           & 56.32          & 30.56          & –              & –              & 73.14          & 41.68          & –     & –        \\ \hline
\multicolumn{15}{c}{Methods for Grounded Situation Recognition} \\  \hline  
ISL \cite{pratt2020grounded}                                         & 38.83          & 30.47          & 18.23          & 22.47          & 7.64           & 65.74          & 50.29          & 28.59          & 36.90          & 11.66          & 72.77          & 37.49          & 52.92 & 15.00    \\
JSL \cite{pratt2020grounded}                                          & 39.60          & 31.18          & 18.85          & 25.03          & 10.16          & 67.71          & 52.06          & 29.73          & 41.25          & 15.07          & 73.53          & 38.32          & 57.50 & 19.29    \\
GSRTR \cite{cho2021grounded}                                        & 41.06          & 32.52          & 19.63          & 26.04          & 10.44          & 69.46          & 53.69          & 30.66          & 42.61          & 15.98          & 74.27          & 39.24          & 58.33 & 20.19    \\
SituFormer \cite{wei2021rethinking}                                  & 44.32 & 35.35 & 22.10 & 29.17 & 13.33 & 71.01 & 55.85 & 33.38 & 45.78 & 19.77 & 76.08 & 42.15 & 61.82 & 24.65    \\
CoFormer \cite{cho2022CoFormer} & 44.41 & 35.87 & 22.47 & 29.37 & 12.94 & 72.98 & 57.58 & 34.09 & 46.70 & 19.06 & 76.17 & 42.11 & 61.15 & 23.09\\
GSRFormer (ours) & \textbf{46.64} & \textbf{37.69} & \textbf{23.58} & \textbf{31.61} & \textbf{14.42} & \textbf{73.43} & \textbf{58.75} & \textbf{35.82} & \textbf{48.42} & \textbf{21.67} & \textbf{78.76} & \textbf{44.71} & \textbf{63.95} & \textbf{25.85}\\
\hline
\end{tabular}}
\vspace{-3mm}
\end{table*}

\vspace{-2mm}
\subsection{Training Objectives}
\label{sec:training-objectives}
We use the same data augmentation and batch training strategy as previous work \cite{cho2021grounded}.
The training details for the encoder and decoder are as follows.

\noindent \textbf{Training of encoder}.
We use the cross-entropy loss function to train the encoder to obtain the pseudo-verb category as,
\begin{equation}
  \mathcal{L}_{\text{verb-e}} = \mathcal L_{\text{CE}}(v^{gt}, \tilde{v}),
\end{equation}
where the ground-truth verb category is denoted as $v^{gt}$ and the predicted pseudo-verb category is $\tilde{v}$.
Note that the first multi-head attention module of the encoder performs gradient back-propagation only when training the encoder, and does not participate in parameter updates when training the decoder.

\noindent \textbf{Training of decoder}.
When training the decoder, we need to optimize the categories of verbs and nouns as well as the bounding boxes of nouns. Following previous work \cite{cho2021grounded,wei2021rethinking}, the losses of the decoder are calculated as,
\begin{equation}
\mathcal{L}_{\text{verb-d}} = \mathcal L_{\text{CE}}(v^{gt}, v),
\end{equation}
\begin{equation}
\mathcal{L}_{nouns} = \sum_{i=1}^{m} \left[\mathcal L_{\text{CE}}(n^{gt}_i,  n_i) + \mathcal{L}_{box}(\mathbf b^{gt}_i, \mathbf b_i)\right],
\end{equation}
where we use the cross-entropy loss function to train the decoder to get the true verb categories $v$.
For the noun loss function,
$n^{gt}_i$ and $\mathbf b^{gt}_i$ denote the ground-truth noun category and bounding box, while $n_i$ and $\mathbf b_i$ are the predicted noun category and bounding box.
$\mathcal{L}_{\text{CE}}$ is the cross-entropy loss for noun classification.
$\mathcal{L}_{box}$ consists of the generalize IoU loss \cite{rezatofighi2019generalized} and the $L1$ regression loss.

\noindent \textbf{Process of inference}.
Similar to previous methods \cite{mallya2017recurrent,pratt2020grounded,wei2021rethinking,cho2021grounded}, GSRFormer also requires inference in the encoder and decoder separately.
Compared to the three-stage SituFormer \cite{wei2021rethinking}, our inference process is more straight.
At inference time, GSRFormer first predicts a pseudo-verb category and then constructs the corresponding semantic roles to learn the representations using the encoder.
Based on the output features from the encoder, a set of similar support verbs is then retrieved in the training set as the input to the decoder.
Finally, the decoder produces the verb and noun predictions.

\begin{table*}[t]
\small
\caption{\small \label{tab:test_set_result} Performance (\%) comparisons of GSRFormer (ours) and baseline methods on SWiG dataset test set.}
\vspace{-3mm}
\resizebox{1.97\columnwidth}{!}{
\begin{tabular}{l|ccccc|ccccc|cccc}
\hline
\multicolumn{1}{c|}{\multirow{2}{*}{Models}} & \multicolumn{5}{c|}{Top-1-Verb}            & \multicolumn{5}{c|}{Top-5-Verb}            & \multicolumn{4}{c}{Ground-Truth-Verb} \\
\multicolumn{1}{c|}{}                        & verb  & value & val-all & grnd  & grnd-all & verb  & value & val-all & grnd  & grnd-all & value  & val-all  & grnd   & grnd-all \\ \hline \hline
\multicolumn{15}{c}{Methods for Grounded Situation Recognition} \\  \hline 
ISL \cite{pratt2020grounded}                                       & 39.36 & 30.09 & 18.62   & 22.73 & 7.72     & 65.51 & 50.16 & 28.47   & 36.60 & 11.56    & 72.42  & 37.10    & 52.19  & 14.58    \\
JSL \cite{pratt2020grounded}                                       & 39.94 & 31.44 & 18.87   & 24.86 & 9.66     & 67.60 & 51.88 & 29.39   & 40.60 & 14.72    & 73.21  & 37.82    & 56.57  & 18.45    \\
GSRTR \cite{cho2021grounded}                                      & 40.63 & 32.15 & 19.28   & 25.49 & 10.10    & 69.81 & 54.13 & 31.01   & 42.50 & 15.88    & 74.11  & 39.00    & 57.45  & 19.67    \\
SituFormer \cite{wei2021rethinking}                                  & 44.20 & 35.24 & 21.86   & 29.22 & 13.41    & 71.21 & 55.75 & 33.27   & 46.00 & 20.10    & 75.85  & 42.13    & 61.89  & 24.89    \\
CoFormer \cite{cho2022CoFormer} & 44.66 & 35.98 & 22.22 & 29.05 & 12.21 & 73.31 & 57.76 & 33.98 & 46.25 & 18.37 & 75.95 & 41.87 & 60.11 & 22.12\\
GSRFormer (ours) & \textbf{46.53} & \textbf{37.48} & \textbf{23.32} & \textbf{31.53} & \textbf{14.23} & \textbf{73.44} & \textbf{58.84} & \textbf{35.82} & \textbf{48.43} & \textbf{21.41} & \textbf{78.81} & \textbf{44.68} & \textbf{63.87} & \textbf{25.35}\\
\hline
\end{tabular}}
\vspace{-3mm}
\end{table*}

\section{Experiment}
\label{sec:experiment}
\subsection{Experimental Settings}
\label{sec:experimental-settings}
\noindent \textbf{Datasets}.
Our experiments are carried out on the challenging SWiG benchmark \cite{pratt2020grounded}.
SWiG dataset builds on the original Situation Recognition (SR) imSitu dataset \cite{yatskar2016situation} by adding bounding box (bbox) annotations for all visible semantic roles (63.9\% of roles have bbox annotations).
Since each image in imSitu is annotated with three verb frames by three annotators, SWiG contains 126,102 images with 504 verbs and 190 semantic role types, and each verb is followed by 1 to 6 semantic roles (3.55 on average).
We followed the official splits to construct the training/validation/testing set with sizes of 75K/25K/25K, respectively.

\noindent \textbf{Evaluation metrics}.
We use the same five evaluation metrics as Pratt \etal~\cite{pratt2020grounded}, including 
1) \textbf{verb}: the accuracy of verb prediction,
2) \textbf{value}: the accuracy of noun prediction for each semantic role,
3) \textbf{val-all}: the accuracy of noun prediction for the whole semantic role set,
4) \textbf{grnd}: the accuracy of noun prediction with correct grounding (bbox) for each semantic role,
5) \textbf{grnd-all}: the accuracy of noun prediction with grounding (bbox) for the whole semantic role set.
Note that we consider a grounding is correct if the IoU between the predicted and ground-truth bbox is above 0.5.
Meanwhile, we report the above metrics in three evaluation settings: 1) \textbf{Top-1-verb}, 2) \textbf{Top-5-verb} and 3) \textbf{Ground-Truth-Verb}, which select verbs based on top-1 prediction, top-5 predictions, and corresponding ground truth, respectively.
If verb predictions are incorrect in the Top-1-verb and Top-5-verb settings, the corresponding semantic role predictions are also considered false.

\vspace{-2mm}
\subsection{Comparisons with State-of-the-Arts}
\label{sec:comparisons-with-state-of-the-arts}
\noindent\textbf{Baseline models}.
Existing SR models can be classified into:
1) \textbf{CRF} \cite{yatskar2016situation}: CRF-based model, 
2) \textbf{CRF+DataAug} \cite{yatskar2017commonly}: CRF-based model with data augmentation,
3) \textbf{VGG+RNN} \cite{mallya2017recurrent}: RNN-based prediction model with VGG backbone,
4) \textbf{FC-Graph} \cite{li2017situation}: GNN-based model with fully connected semantic roles,
5) \textbf{CAQ} \cite{cooray2020attention}: Query-based model with top-down attention,
6) \textbf{Kernel-Graph} \cite{suhail2019mixture}: GNN-based model with mixture kernel attention.
Correspondingly, existing GSR models can be divided as:
1) \textbf{ISL} \cite{pratt2020grounded}: RNN-based method has independent semantic role values and grounding predictions.
2) \textbf{JSL} \cite{pratt2020grounded}:
RNN-based methods jointly predict semantic role values and their basis.
3) \textbf{GSRTR} \cite{cho2021grounded}:
Transformer two-stage model has both a verb predictor and semantic role detector.
4) \textbf{SituFormer} \cite{wei2021rethinking}:
Transformer three-stage model consists of a coarse-to-fine verb predictor and a semantic role detector.
4) \textbf{CoFormer} \cite{cho2022CoFormer}:
Transformer three-stage model exploits the semantic relations between the verb and noun roles to improve results.

\noindent \textbf{Results under Ground-Truth-Verb setting}.
The Ground-Truth-Verb setting evaluates whether the system can understand events in a human-cognitive manner.
The numerical value of this setting describes how well the machine predictions match the human annotations.
The experimental results are shown in Table \ref{tab:dev_set_result} and Table \ref{tab:test_set_result}.
In general, our GSRFormer outperforms other state-of-the-art methods.
Compared with SituFormer\cite{wei2021rethinking}, which also adopts the transformer structure, GSRFormer improves the accuracy of noun prediction in single (value) and all semantic roles (val-all) by 2.96\% and 2.55\%, respectively.
Furthermore, GSRFomer achieves similar improvements under the vision grounding setting (i.e., grnd and grnd-all), which shows that GSRFormer can effectively learn visual information from natural scenes.

\begin{table}[]
\centering
\small
\caption{\small Effectiveness of each component of GSRFormer.}
\label{tab:ablation-component}
\vspace{-3mm}
\resizebox{0.90\columnwidth}{!}
{
\begin{tabular}{l|ccccc}
\hline 
Components & verb           & value          & val-al         & grnd           & grnd-all       \\
\hline \hline
w/o Encoder-1st   & 35.30  & 25.44  & 15.69 & 21.27  & 7.60   \\
Gains ($\Delta$)  & -11.23 & -12.04 & -7.63 & -10.26 &\textbf{ -6.63  }\\
w/o Encoder-2nd   & 32.84  & 24.21  & 14.60 & 20.91  & 7.66   \\
Gains ($\Delta$)  & \textbf{-13.69} &\textbf{ -13.27 }&\textbf{ -8.72} &\textbf{ -10.62 }& -6.57  \\ \hline
w/o Decoder       & 34.94  & 25.08  & 14.28 & 20.99  & 7.79   \\
Gains ($\Delta$)  & -11.59 & -12.40 & -9.04 & -10.54 & -6.44   \\ \hline
w/o Iteration     & 39.10  & 31.30  & 19.11 & 26.18  & 11.92   \\
Gains ($\Delta$)  & -7.43  & -6.18  & -4.21 & -5.35  & -2.31   \\
w/o Alternate     & 35.81  & 27.87  & 16.39 & 22.09  & 9.17    \\
Gains ($\Delta$)  & \textbf{-10.72} &\textbf{ -9.61}  &\textbf{ -6.93} & -9.44  & -5.06   \\ 
w/o Message & 38.06  & 29.87  & 17.69 & 21.83  & 7.77          \\
Gains ($\Delta$) & -8.47  & -7.61  & -5.63 &\textbf{ -9.70}  & \textbf{-6.46 }   \\ \hline
GSR-Former (ours) & 46.53  & 37.48  & 23.32 & 31.53  & 14.23   \\
\hline 
\end{tabular}}
\vspace{-6mm}
\end{table}

\noindent \textbf{Results under Top-N-Verb settings}.
We use the Top-N-Verb setting to evaluate the accuracy of predicting verb categories.
The results in Table \ref{tab:dev_set_result} and Table \ref{tab:test_set_result} show that our GSRFormer outperforms other state-of-the-art methods.
It is well known that SituFormer \cite{wei2021rethinking} adopts a three-stage framework to improve the accuracy of verb prediction. 
Compared to SituFormer, GSRFormer can further push the verb prediction accuracy by 2.33\% (absolute) under the Top-1-Verb setting.
In addition, GSRFormer also achieves a more splendid improvement on SituFormer in single noun prediction (2.24\% on value) under the Top-1-Verb setting.
Unlike SituFormer, which only uses nouns to improve verbs, our GSRFormer adopts a bidirectional refinement strategy to iterative optimize the results.
The 2.24\% increase in noun prediction accuracy fully reveals the effectiveness of our proposed alternative semantic refinement.

\begin{table}[]
\centering
\small
\caption{\small Effects of adopting two opposite refinement orders.}
\vspace{-3mm}
\label{tab:refinement-order}
\resizebox{0.82\columnwidth}{!}
{
\begin{tabular}{l|ccccc}
\hline
\multicolumn{1}{c|}{\multirow{2}{*}{Order}} & \multicolumn{5}{c}{Ground-Truth-Verb}      \\
\multicolumn{1}{c|}{}                      & verb  & value & val-all & grnd  & grnd-all \\ \hline \hline
Refine-Verb-First                          &    -   & 75.45 & 42.29   & 61.61 & 24.72    \\
Refine-Noun-First                          &   -    & \textbf{78.81} & \textbf{44.68}   & \textbf{63.87} & \textbf{25.35}    \\ \hline
                                           & \multicolumn{5}{c}{Top-1-Verb}             \\ \hline
Refine-Verb-First                          & 45.23 & 36.26 & 22.57   & 31.12 & 13.65    \\
Refine-Noun-First                          & \textbf{46.53} & \textbf{37.48} & \textbf{23.32}   & \textbf{31.53} & \textbf{14.23} \\
\hline
\end{tabular}}
\vspace{-4mm}
\end{table}

\begin{table}[]
\centering
\small
\caption{\small Effects of utilizing different aggregate functions.}
\label{tab:ablation-aggregate-functions}
\vspace{-3mm}
\resizebox{0.82\columnwidth}{!}
{
\begin{tabular}{l|cccc}
\hline 
Aggregate Functions             & value          & val-all        & grnd           & grnd-all       \\
\hline \hline
Element-wise Sum         & 75.64          & 42.79          & 61.31          & 24.38          \\
Max-Pooling        & 76.30          & 42.94          & 61.85          & 24.23          \\
Aggregated Message \cite{kipf2016semi} & 77.79          & 43.58          & 62.30          & 24.76          \\
Gated Function \cite{li2016gated}    & \textbf{78.81} & \textbf{44.68} & \textbf{63.87} & \textbf{25.35}\\
\hline
\end{tabular}}
\vspace{-6mm}
\end{table}

\subsection{Ablation Studies}

\begin{figure*}[th]
  \centering
  \small
  \includegraphics[width=0.85\linewidth]{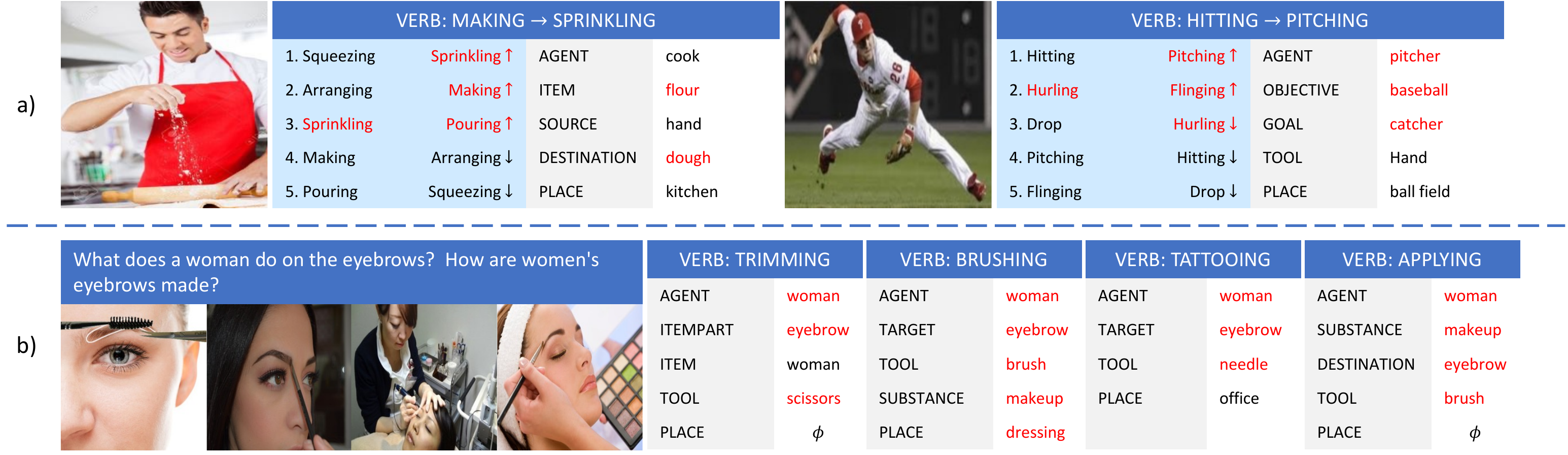}
  \vspace{-3mm}
  \caption{\small (a) Comparison of GSRFormer and GSRTR \cite{cho2021grounded}, where GSRFormer predicts more correct verbs and nouns under Top-5-Verb setting.
  (b) An example demonstrates the application of GSRFormer, which can serve cross-modal semantic question answering.}
  \label{fig:experiment-example}
  \vspace{-2mm}
\end{figure*}

\begin{figure*}[th]
  \centering
  \small
  \includegraphics[width=0.85\linewidth]{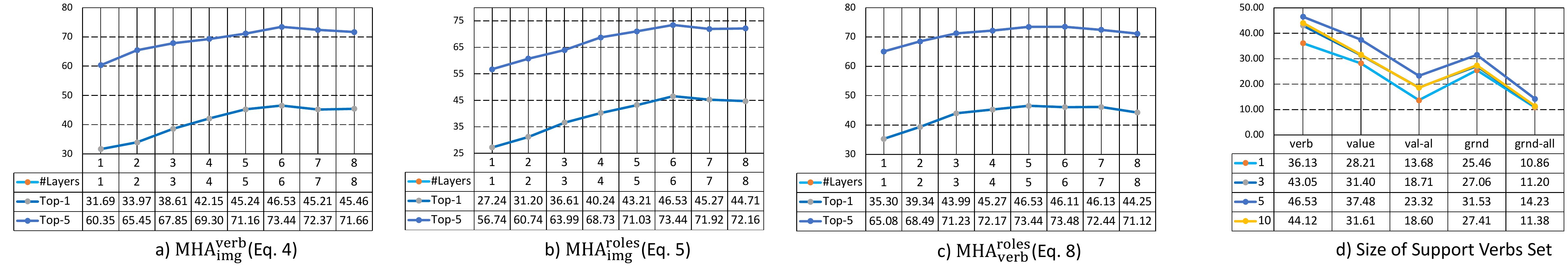}
  \vspace{-5mm}
  \caption{\small (a)-(c) show the verb prediction accuracy under the Top-1-Verb and Top-5-Verb settings.
Here we verify the effect of different numbers of stacked layers on three multi-head attention structures.
(d) reveal the effect of the size of the support verbs under the Top-1-Verb setting.}
  \vspace{-2mm}
  \label{fig:ablation-study}
\end{figure*}

\noindent \textbf{Effectiveness of encoder}.
We perform ablation studies on two multi-head attention modules of the encoder (denoted as Encoder-1st and Encoder-2nd), as shown in Table \ref{tab:ablation-component}.
Experimental results show that both structures significantly improve performance (over 10\% absolute improvement).
This fully demonstrates the effectiveness of the encoder module.
Encoder-1st is valuable in understanding visual grounding details (comparable on grnd-all metrics), while Encoder-2nd is more conducive to mining verbs and semantic roles.

\noindent \textbf{Effectiveness of decoder}.
We validate the entire decoder module.
By removing the decoder, the modified model resembles a simple two-stage approach \cite{cho2021grounded}.
The results in Table \ref{tab:ablation-component} show that the absolute improvement is also over 10\%, which fully demonstrates the effectiveness of the decoder module.
Below we will analyze each part of the decoder in detail.

\noindent \textbf{Effectiveness of iterative refinement}.
We verify iterative refinement.
Without iterative refinement (i.e., only 1 transformer layer in the decoder instead of 5), the performance (2\%-8\% drop) will be similar to traditional RNN-based methods (e.g., JSL\cite { pratt2020grounded}).
It hints that the real improvement to the transformer structure is the ability to make iterative improvements.

\noindent \textbf{Effectiveness of alternate optimization}.
We verify alternate optimization.
By removing this, we mean to update the verb with support verbs, and update nouns with other nouns.
We are surprised to find that alternate optimization resulted in more performance drop than iterative refinement, except for the visual relevant grnd-all metrics.
This fully illustrates that the core of iterative refinement is to exploit the semantic relationship between verbs and nouns (roles) in both directions.

\noindent \textbf{Effectiveness of message computation}.
We compare with that removes the message passing mechanism, i.e., directly using features of transformer encoder to learn the semantic relations.
As shown in Table \ref{tab:ablation-component}, message passing not only effectively uses semantic relations to improve verb and noun prediction, but also learns the visual grounding information (grnd and grnd-all metrics).

\noindent \textbf{Effects of refinement order}.
Table \ref{tab:refinement-order} shows that refining the noun first gives better results.
This is the exact opposite of the previous work \cite{mallya2017recurrent, pratt2020grounded, cho2021grounded}.
We speculate that this is because the set of noun entities is more relevant to the visual groundings, so predicting nouns first leads to fewer errors than verbs.
For example, humans naturally recognize noun entities as evidence for judgments when stating verbs.
This inspires us to prioritize noun entities.

\noindent \textbf{Effects of aggregate functions}.
We test four aggregate functions.
Table \ref{tab:ablation-aggregate-functions} shows that Gated Function \cite{li2016gated} achieves the best results and Element-Wise Sum does not reach satisfactory results.
This contrast points out that aggregation functions should strive to highlight semantically relevant features while ignoring unnecessary noise.
In future work, we will explore more fusion strategies.

\noindent \textbf{Effects of the number of stacked layers}.
We validate the effect of stacking layers in three multi-head attention structures.
As shown in Figure \ref{fig:ablation-study}, the best performance is achieved when six layers are stacked in the encoder and five layers in the decoder.
As the number of stacked layers increases, the performance first increases and then decreases.
We attribute this performance change to the noise of stacking too many layers.

\noindent \textbf{Effects of the size of support verb set}.
As shown in Figure \ref{fig:ablation-study} (d), the best performance is when the support set size is 5.
The performance drops when the size is 1, indicating that similar verbs can support semantic understanding.
Moreover, when the size is 10, the performance does not further improve, which means that expanding support verbs also raises noise.

\vspace{-4mm}
\subsection{Visualization and Application}
We visualize the results of GSRFormer in Figure \ref{fig:experiment-example} (a).
With the help of alternate semantic refinement, GSRFormer predicts some almost indistinguishable action verbs and semantic roles (see red font).
We also show an example of the semantic question-answering application in Figure \ref{fig:experiment-example} (b).
For example, when we ask questions like "\textit{How are women's eyebrows made}?", GSRFormer can not only utilize the generated verb-frame to understand the question, but also provide structured answers with rich image facts.

\section{Conclusion}
In this paper, we first reveal the problems of existing frameworks and point out that the use of semantic relations is the root of the GSR task.
To this end, we propose GSRFormer, a two-stage transformer framework that utilizes bidirectional semantic relations to iteratively refine predictions of verb and noun entities.
Experimental results on SWiG dataset show that it outperforms other state-of-the-art methods.
In the future, we will further explore to explain the semantic structure of GSRFormer and extend it to other semantic analysis tasks.

\begin{acks}
{This work was supported by the Air Force Research Laboratory under agreement number~FA8750-19-2-0200;
the Defense Advanced Research Projects Agency~(DARPA) grants funded under the GAILA program~(award HR00111990063);
the Defense Advanced Research Projects Agency~(DARPA) grants funded under the AIDA program~(FA8750-18-20018).

The U.S. Government is authorized to reproduce and distribute reprints for Governmental purposes notwithstanding any copyright notation thereon. The views and conclusions contained herein are those of the authors and should not be interpreted as necessarily representing the official policies or endorsements, either expressed or implied, of the Air Force Research Laboratory or the U.S. Government.}
\end{acks}

\clearpage
\bibliographystyle{ACM-Reference-Format}
\balance
\bibliography{sample-base}

\appendix
\section{Implementation Details}
\label{sec:implementation-details}
\noindent \textbf{Network structure}.
We use ImageNet pre-trained ResNet-50 as the backbone.
To facilitate computation, we use 1×1 convolution to compress the dimension of image features to 512, which is consistent with the hidden dimension of each semantic role query and verb token.
To further speed up inference, the dimension of verb embedding and role embedding is 256.
Correspondingly, we use learnable 2D embeddings for positional encodings, which have the same dimensions as sequences of visual and verb tokens.
The number of heads for all multi-head attention blocks is 8.
The sizes of the hidden dimensions of the four structures are 2048, 1024, 1024, and 1024, and the dropout rates are 0.15, 0.3, 0.3, and 0.2, respectively.
The bounding box regressor is three fully connected layers with ReLU activation function and 1024 hidden dimensions, using a dropout rate of 0.2.
Label smoothing regularization \cite{szegedy2016rethinking} is used for target verb and noun labels with label smoothing factors of 0.3 and 0.2, respectively.
We set the support verb set size to 5.
The number of stacked layers of the encoder is set to 6, and the number of stacked layers of the decoder is set to 5.

\begin{figure*}[th]
\small
\centering
\includegraphics[width=1.0\linewidth]{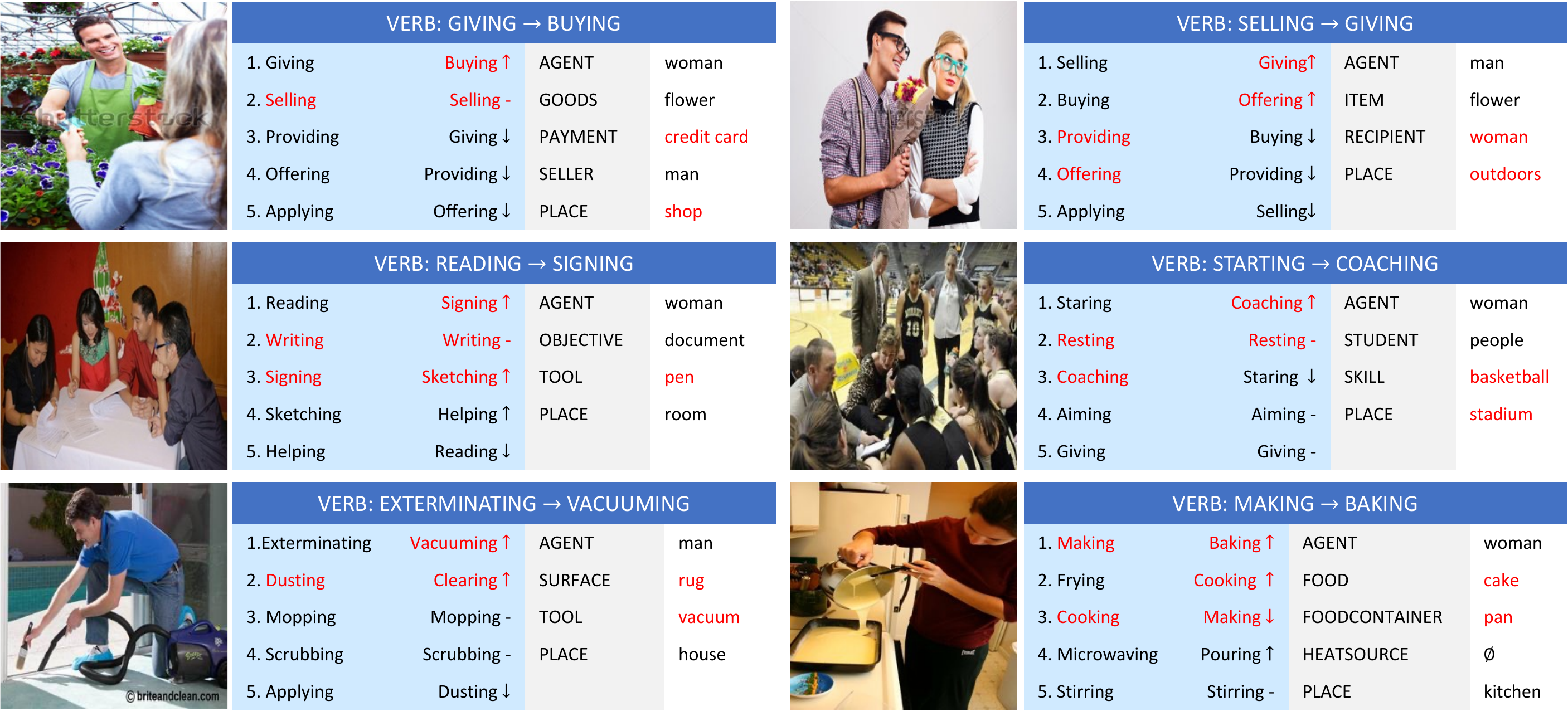}
\vspace{-5mm}
\caption{\small Comparison between GSRFormer and GSRTR \cite{cho2021grounded}.
The first two columns compare the verbs detected by GSRTR and GSRFormer, respectively.
The last two columns are the semantic roles predicted by GSRFormer.
With the help of an iterative refinement mechanism, GSRFormer predicts more correct verbs and nouns (marked in red font) under the Top-5-Verb setting.
These examples also clearly illustrate the importance of semantic relations for GSR tasks, i.e., bidirectional semantic ties can mutually refine the predictions of verbs and nouns.}
\vspace{-1mm}
\label{fig:compare-result-sm}
\end{figure*}

\begin{figure*}[th]
\small
\centering
\includegraphics[width=1.0\linewidth]{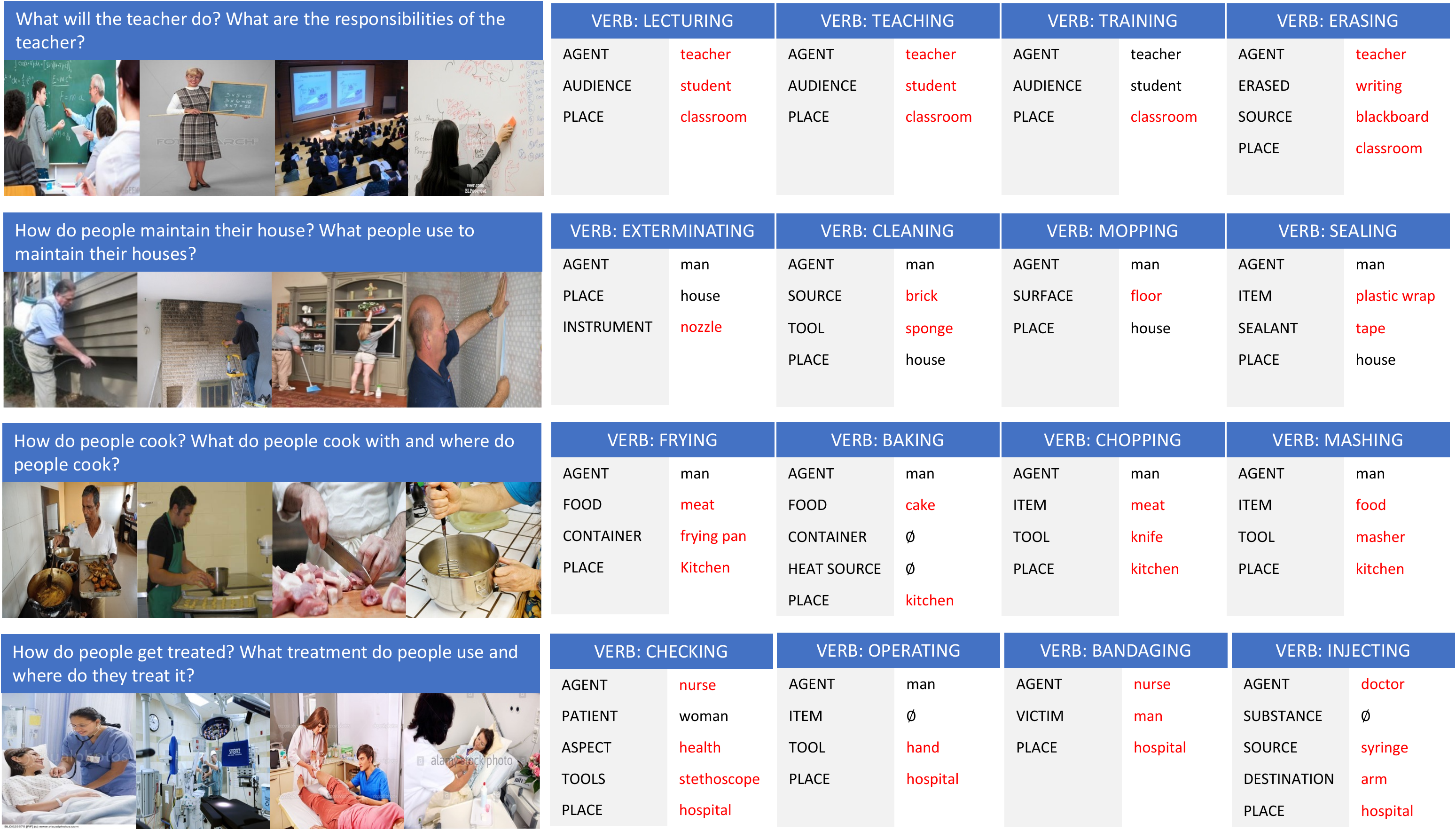}
\vspace{-5mm}
\caption{\small 
Application example.
GSFormer can serve cross-modal semantic question answering and reasoning on the basis of human event understanding.
For example, when questioned "\textit{How do people cook? What do people cook with, and where do people cook}?", GSRFormer can list cooking procedures and steps.
Compared to image captioning and scene graphs, GSRFormer can not only utilize the generated structured verb-frame to apprehend the questions, but also provide answers with intuitive image facts to help users understand.}
\label{fig:application-sm}
\end{figure*}

\noindent \textbf{Training details}.
Although using the data augmentation strategies similar to DETR \cite{carion2020end} can improve the experimental results, for a fair comparison, we employ the same data augmentation procedure as previous work \cite{cho2021grounded}.
Specifically, Random Color Jittering, Random Grayscale Scaling, Random Scaling, and Random Horizontal Flipping are employed. The Hue, Saturation, and Lightness Scales in random Color Jittering are set to 0.1. Random Grayscale Scaling is set to 0.3. Random Scaling is set to 0.5, 0.75, and 1.0. The probability of Random Horizontal Flip is set to 0.5.
The number of semantic roles varies from 1 to 6, depending on the verb category.
To speed up training, inspired by previous work \cite{cho2021grounded}, we utilize zero padding for each output of the noun prediction branch to ensure batch training.
Because there are as many semantic role queries as semantic roles, we directly ignore the padding output in the loss computation.
When training the decoder, we need to alternately compute verb and noun losses separately.
Since there are three nouns per semantic role, the final noun loss is the sum of the three noun losses.
In addition, we also illustrate more visualization comparisons and application examples, as shown in Figure \ref{fig:compare-result-sm}-\ref{fig:application-sm}.

\vspace{-2mm}

\end{document}